\RequirePackage{fix-cm}
\documentclass[twocolumn]{svjour3_temp}          
\smartqed  
\def\eg{\emph{e.g.}}

\def\ie{\emph{i.e.}}

\def\etal{\emph{et al.}}

\newcommand{\mname}{Noisy-LSTM}
\newcommand{\Mname}{Noisy-LSTM}
\usepackage{multirow}
\usepackage{amsmath}
\usepackage{booktabs} 
\usepackage{color}
\usepackage{graphicx}
\begin{document}

\title{Noisy-LSTM: Improving Temporal Awareness for Video Semantic Segmentation
}


\author{Bowen Wang         \and
        Liangzhi Li \and
        Yuta Nakashima \and
        Ryo Kawasaki\and
        Hajime Nagahara\and
        Yasushi Yagi
}


\institute{Bowen Wang \at
              \email{bowen.wang@is.ids.osaka-u.ac.jp}           
           \and
           Liangzhi Li \at
            \email{li@ids.osaka-u.ac.jp}
            \and
           Yuta Nakashima \at
            \email{n-yuta@ids.osaka-u.ac.jp}
            \and
           Ryo Kawasaki\at
            \email{ryo.kawasaki@ophthal.med.osaka-u.ac.jp}
            \and
           Hajime Nagahara\at
           \email{nagahara@ids.osaka-u.ac.jp}
           \and
           Yasushi Yagi\at
           \email{yagi@am.sanken.osaka-u.ac.jp}
}


\maketitle

\begin{abstract}
Semantic video segmentation is a key challenge for various applications. This paper presents a new model named \mname{}, which is trainable in an end-to-end manner, with convolutional LSTMs (ConvLSTMs) to leverage the temporal coherency in video frames. We also present a simple yet effective training strategy, which replaces a frame in video sequence with noises. This strategy spoils the temporal coherency in video frames during training and thus makes the temporal links in ConvLSTMs unreliable, which may consequently improve feature extraction from video frames, as well as serve as a regularizer to avoid overfitting, without requiring extra data annotation or computational costs. Experimental results demonstrate that the proposed model can achieve state-of-the-art performances in both the CityScapes and EndoVis2018 datasets.  Code is available at https://github.com/wbw520/NoisyLSTM.
\keywords{Semantic Segmentation \and Noise \and Temporal Awareness}
\end{abstract}

\section{Introduction}
The ever-increasing importance of video semantic segmentation has attracted a fast-growing attention from an extensive number of computer vision researchers. Due to the rapid development of convolutional neural networks (CNNs) \cite{yu2015multi,shin2016deep}, it is fair to say that the performance of video semantic segmentation has been dramatically improved.
A simple yet effective approach is to treat video frames as independent images and use image segmentation models for each frame. This approach can benefit from many well-developed image segmentation models \cite{chen2017rethinking,fu2019dual,zhao2017pyramid} and the large number of available training datasets \cite{everingham2015pascal,lin2014microsoft}. 

\begin{figure*}
\centering 
\includegraphics[width=0.95\textwidth]{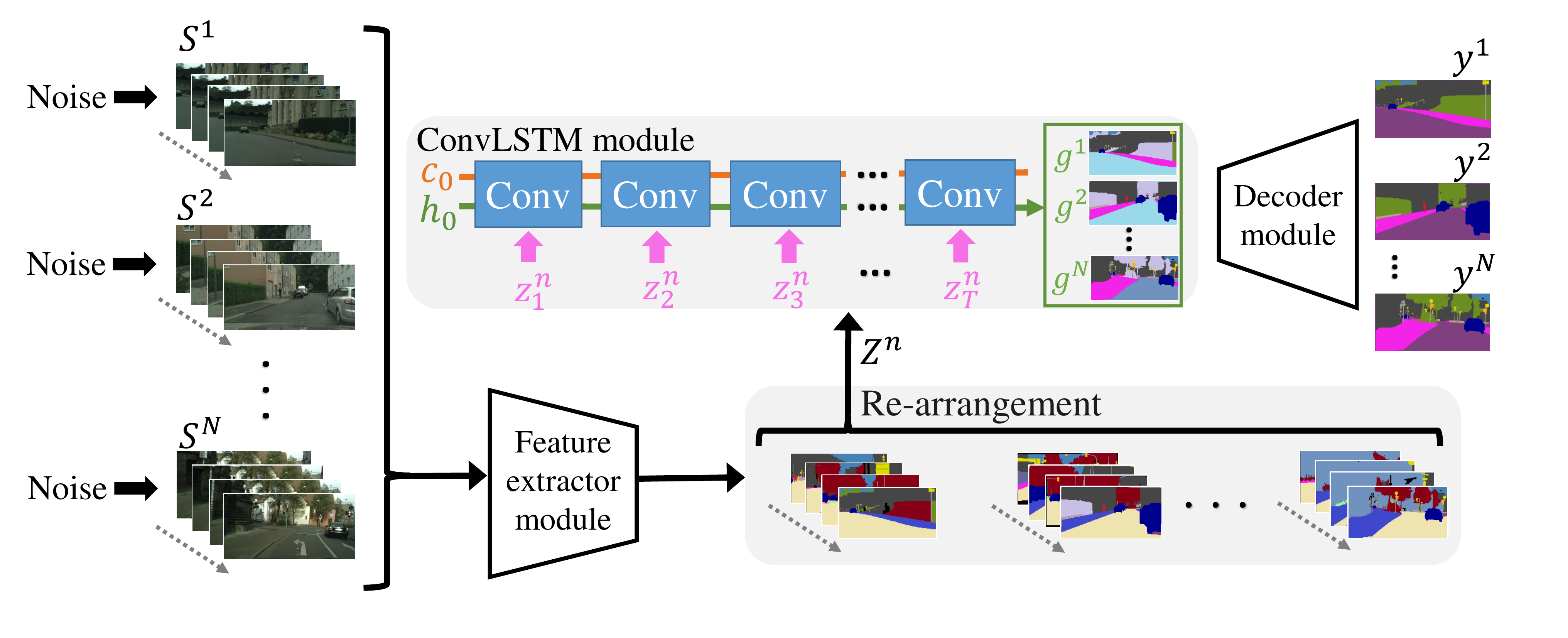}
\caption{Overview of the proposed \mname{} model for video semantic segmentation.}
\label{fig:overview} 
\end{figure*}

However, these methods usually suffer from some segmentation errors like inaccurate object boundaries, incomplete regions that only cover parts of certain objects, and over-complete regions that cover neighboring objects. Due to the deteriorated imaging and color quality caused by video capturing and encoding, these segmentation errors happen much more frequently in video semantic segmentation tasks. 
An important observation is that these errors only exist in some frames, while other frames, including adjacent ones, may still get accurate predictions. 

Based on this observation, researchers have developed new models dedicated for video semantic segmentation that utilize the temporal coherency. There are some works that use optical flow \cite{gadde2017semantic,nilsson2018semantic,jin2017predicting},
whereas the computation of optical flow itself is a non-trivial problem that depends much on the motion dynamics in adjacent frames. It is hard to design a robust and accurate method for estimating optical flow for a variety of videos. 

Another possible way to leverage the temporal coherency is to introduce temporal structure in models. One pioneering approach is to use conditional random fields (CRFs) on top of a model for a single image with corresponding variables connected in the temporal dimension \cite{xu2012streaming,kundu2016feature}. However, their CRFs have no access to internal representations in the CNNs, which may spoil their potential to improve the segmentation results. 
Recurrent neural networks (RNNs) provide further flexibility, and there have been a series of works  \cite{valipour2017recurrent,emre2017semantic,rochan2018future,pfeuffer2019semantic}. 
They used recurrent networks to extract relationship information of adjacent frames for current frames prediction. However, additional links in the temporal dimension introduce more model parameters to be trained and may require more training data. Especially, most RNN-based models need a large number of labeled data for training, which may not always available for many applications.

Data augment is a possible way to fix these kinds of problems. Recent techniques for training neural networks sometimes use noises. For example, dropout and its related techniques \cite{labach2019survey} inject noises into latent representations to regularize training. Some methods add noises even to input images as data augmentation \cite{nazare2017deep}. Xie \etal \cite{xie2019self} proposed to use unlabeled data, which served as noise for training, in a teacher-student framework. The experimental results in these works demonstrate that using noises in training is an easy yet effective way to improve the performance.

In this paper, we propose a new method named \mname{}, which uses convolutional LSTM (ConvLSTM \cite{xingjian2015convolutional}) to facilitate the temporal continuity to improve video semantic segmentation tasks. Inspired by \cite{xie2019self}, we adopts a new noisy-training strategy to further improve its ability to utilize the temporal coherency. As shown in Fig.~\ref{fig:overview}, the \mname{} model is based on a feature extractor and extended with ConvLSTM to leverage temporal coherency. \Mname{} can be applied to all common semantic segmentation models. \mname{} uses multiple sequences as input, 
into which random tensors are added. All frames in these sequences are compiled into a single batch and are fed into a shared CNN, in which batch normalization stabilizes the training process. Resulting feature maps are re-arranged into the original sequences, and each of them goes through CovnLSTM module to make use of their temporal dynamics for prediction. Ultimately, the decoder generates semantic segmentation results. 

Our main contribution is three-fold:
\begin{itemize}
	\item We develop a video segmentation method that makes use of the temporal coherence in video frames with ConvLSTM.
	\item We also enhance the model's temporal awareness by using a noisy-training strategy. Without any extra data annotation or computation costs, our strategy regularizes the training. This strategy can be also viewed as a way to control reliability of temporal connections. We can apply this strategy to other models for improving their semantic segmentation performance.
	\item We experimentally demonstrate that our model trained with the noisy-training strategy outperformed or is comparable to the state-of-the-art models over the Cityscapes and EndoVis2018 datasets.  
\end{itemize}

\section{Related work}
In this section, we will here briefly review the representative literature.
\paragraph{Time Sequence Semantic Segmentation}Most approaches are designed only for image segmentation and not for video task. It means the temporal coherency of the video is not considered and each frame of a video sequence is predicted independently. A common approach to deal with the temporal coherency is to use RNN-based structures like Long Short Term Memory (LSTM) networks \cite{hochreiter1997long}. On top of fully convolutional networks (FCNs) \cite{long2015fully}, Valipour \etal introduced the recurrent fully convolutional network (RFCN) \cite{valipour2017recurrent}. They added a recurrent unit between the encoder and the decoder in a FCN, and achieved a better performance on the SegTrack, Davis, and Moving MNIST datasets. Yurdakul \etal \cite{emre2017semantic} evaluated different kinds of RNN-based structures, such as ConvRNN, ConvGRU, and ConvLSTM, on the virtual KITTI dataset \cite{gaidon2016virtual}, and conculuded that ConvLSTM had the best performance. Nilsson and Sminchisescu \cite{nilsson2018semantic} used optical flow to represent changes between adjacent frames and applied the ConvGRU structure to encode temporal continuity. In addition, they used unlabeled frames to further improve the prediction performance. Rochan \etal \cite{rochan2018future} adopted bidirectional ConvLSTM for future frame prediction. They added the ConvLSTM structure between each layer in the encoder and decoder, merging the temporaly adjacent feature maps to predict the target frame. Pfeuffer \etal \cite{pfeuffer2019semantic} applied ConvLSTM at different positions of some state-of-the-art models and demonstrated that ConvLSTM worked well with most positions. 

\paragraph{Training with Noises} For the training of deep models, insufficient training data is a crucial issue that causes overfitting. In order to avoid this issue, various ways to use noises during training have been proposed. Dropout \cite{hinton2012improving} is one of them, adding noises to latent representations in neural networks. There are some variants of dropout \cite{labach2019survey}. Data augmentation by adding noises is also considered \cite{nazare2017deep}, where the equivalence between data augmentation by noises and dropout is pointed out \cite{noh2017regularizing}. Recently, using unlabeled data to improve the model performance is proved possible. Xie \etal \cite{xie2019self} proposed a self-training method named Noisy Student to improve the classification performance on the ImageNet dataset. $300$M unlabeled images, many of which were from different domains, were used to enhance the feature extraction ability of the student model. They applied the teacher-student approach in semantic segmentation tasks for images and presented a new model compression method that can result in models with a good performance while having a much smaller parameter size.

In this paper, we also use unlabeled data to improve the segmentation performance, one of the biggest differences is that our strategy does not require a dual-network structure like teacher-student, as well as the temporarily-generated labels, or the iterated-training process, which cost more time and resources.
We borrow the insight that adding noises in training enhances the feature extraction capability of a model, and propose to add noises in temporal sequences. With this strategy, we expect that the model is robust to occasional and rare changes in frames, which cannot be handled only by a ConvLSTM-based network. 

\section{Methodology}

\subsection{Our Model}\label{model}

As shown in Fig.~\ref{fig:overview}, the proposed model mainly consists of three components: feature extraction module, ConvLSTM module, and decoder module. It takes multiple sequences in a batch $S=\{S^n|n=1,\dots,N\}$ as input, where $N$ is the batch size, and produces a single segmentation result for each sequence as output. Input sequence $S^n = \{s^n_{t}|t=1,\dots,T\}$ contains $T$ frames, where the last frame $s^n_T$ is the target frame for which our model produces segmentation result $y^n$, and other frames $s^n_t$ for $t \not= T$ contextualize $s^n_T$. $T$ is fixed in our implementation, and thus all input sequences have the same length $T$. Note that $s^n_{t}$ and $s^{n}_{t+1}$ are not necessarily consecutive in the original video sequence, but they can be frames separated by a fixed number of frames.

For PSPNet based model, our feature extraction module adopt ResNet-101 \cite{he2016deep} as the backbone network. We replace the last two convolution layers of ResNet-101 with dilated convolutions \cite{yu2015multi} of size $3 \times 3$, rate of 2 and 4 to enlarge the receptive field and remove the fully-connected layers in original ResNet-101. Batch normalization (BN) is of great value for training deep models  \cite{pfeuffer2019semantic}, but it requires diversity in an input batch; otherwise, it may cause severe performance degradation \cite{singh2019filter}. This is a serious problem for models that deal with temporal sequences, because they only input the frames from same video sequences, which may not offer enough diversity. This can be the main reason why most LSTM-based video segmentation models \cite{nilsson2018semantic,pfeuffer2019semantic} do not have BN layers. To address this, in the training stage, we sample target frames $s_T^n$ randomly from all frames in the training set and then aggregate context frames for each target frame to form sequence $S^n$. Also, the feature extraction module does not aware of the sequence structure, \ie, it flattens all sequences into a set of $T \times N$ frames, so that we can easily apply BN. We denote feature map obtained from $s_t^n$, which is the output of the second dilated convolution layer, by $z^n_t$.

\begin{figure*}[t]
\centering 
\includegraphics[width=1\textwidth]{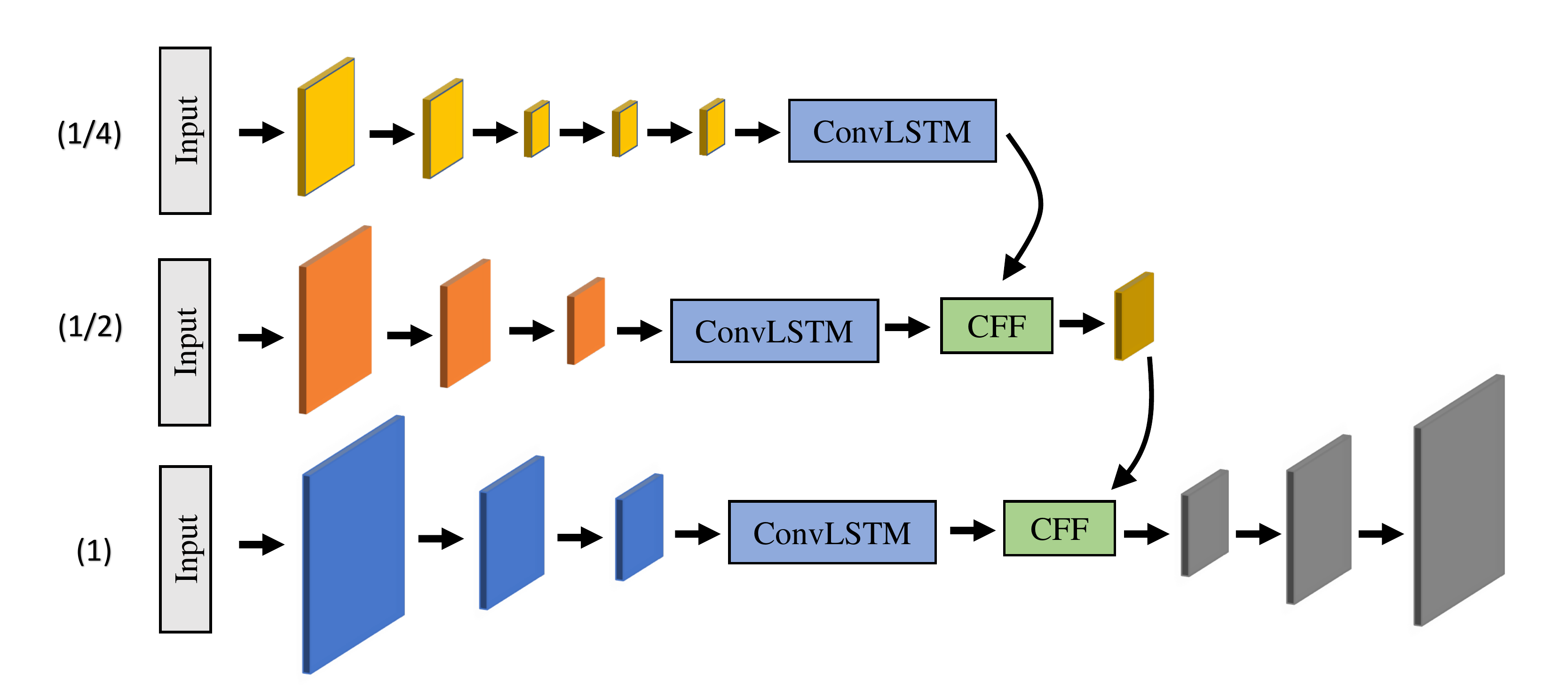}
\caption{ICNet-based \mname{}. We add ConvLSTM module directly after the feature extractor of each branch and the output features are aggregated by the CFF module.}
\label{fig:IC_based} 
\end{figure*}

\begin{figure*}[t]
\centering 
\includegraphics[width=1\textwidth]{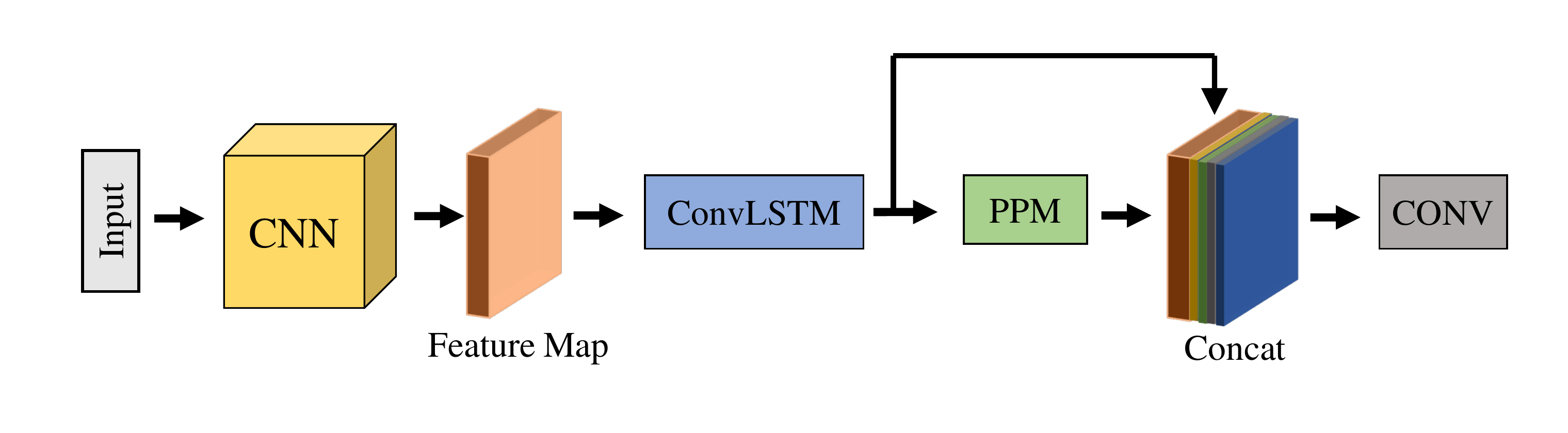}
\caption{PSPNet-based \mname{}. We add ConvLSTM after the CNN feature extractor and the output features will go through the PPM for final prediction.}
\label{fig:PSP_based} 
\end{figure*}


The ConvLSTM module to encode the temporal sequence into a single feature map, which will be detailed in the next section. The previous work \cite{pfeuffer2019semantic} proved that ConvLSTM can be used for various stages (\ie, layers) in various model architectures. We put the ConvLSTM between the feature extraction and decoder modules. The output of ConvLSTM module can be represented by
\begin{equation}
\begin{aligned}
g^n=\operatorname{ConvLSTM}\left(Z^n\right),
\end{aligned}
\end{equation}
where $Z^n = \{z_t^n | t=1,\dots,T\}$. 

Finally, the decoder module takes the outputs $g^n$ from the ConvLSTM module and produces semantic segmentation result $y^n$ for target frame $s_T^n$ of input sequence $S^n$. 

In this paper, we apply \mname{} to ICNet \cite{zhao2018icnet} and PSPNet \cite{zhao2017pyramid} and the model structures are respectively shown in Fig.~\ref{fig:IC_based} and Fig.~\ref{fig:PSP_based}.
For ICNet-based \mname{}, we directly add ConvLSTM module at the end of each branch and the output features are aggregated by the cascade feature fusion (CFF) module.
In PSPNet-based \mname{}, the ConvLSTM module is placed between the underlying CNN model and the decoder module which consists of a pyramid pooling module (PPM), two convolutional layers, and an upsampling layer. 

In what follows, we detail our network design to encode the temporal dependency through ConvLSTM and enhancement of temporal awareness by the noisy-training strategy. 

\begin{figure*}[t]
\centering 
\includegraphics[width=1\textwidth]{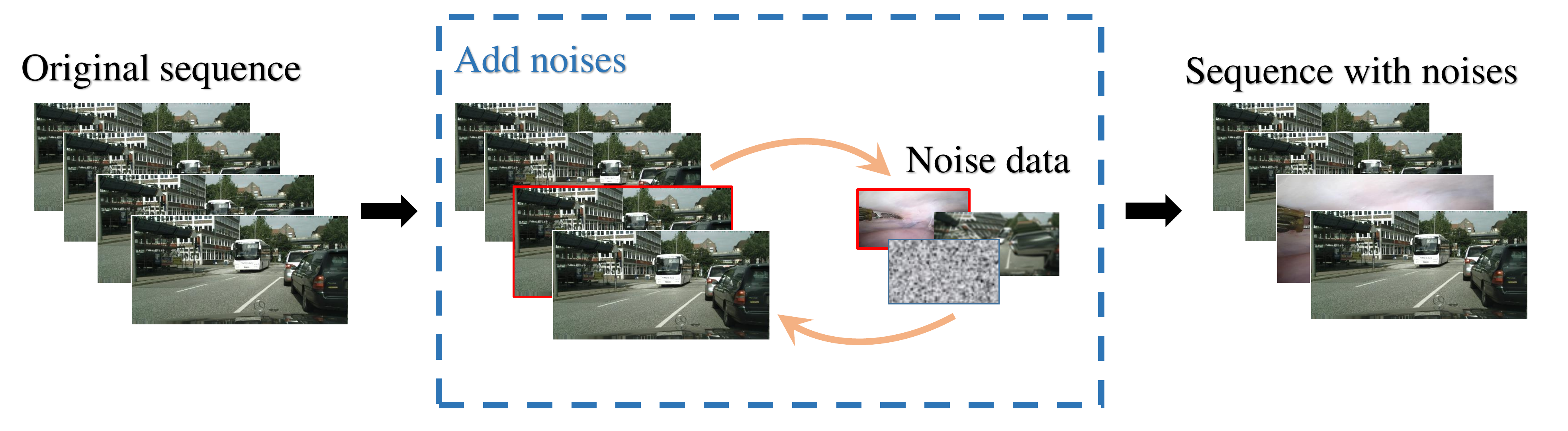}
\caption{A noisy-training strategy introduces noises in the time domain during the training process, by replacing some frames in the sequence with random images.
}
\label{fig:noise_pic} 
\end{figure*}

\subsection{Encoding Temporal Dependency}\label{Conv}
It is proved that ConvLSTM is a powerful tool for capturing the spatio-temporal dependency, which is important for semantic segmentation in video \cite{pfeuffer2019semantic}. The LSTM cells can learn how to handle information from precedent frames  during training and is able to memorize information over a certain period.  In contrast to LSTMs for fully-connected layers \cite{graves2013generating}, ConvLSTMs use as latent state a convolutional layer, which is more suitable for vision tasks. We use a single layer ConvLSTM and set the kernel size to $3 \times 3$. The segmentation result for the target frame ($t=T$) is given based on its own and the precedent ($t=1, \dots, T-1$) frames' feature maps.

As shown in Fig.~\ref{fig:overview}, the feature map from each input frame is sequentially fed into the ConvLSTM layer to get the feature map based on which the segmentation result for the target frame are computed. Formally, from feature map $z_{t}$ for the $t$-th frame in input sequence $S$ (we omit the superscript $n$ for notation simplicity), $g$ is computed as the last latent state of the ConvLSTM layer as follows: 
\begin{equation}\label{ConvLSTM}
\begin{aligned}
i_{t}=& \sigma(W_\text{i}*z_{t} + V_\text{i}*h_{t-1} + U_\text{i} \otimes c_{t-1} + b_\text{i})\\
f_{t}=& \sigma(W_\text{f}*z_{t} + V_\text{f}*h_{t-1} + U_\text{f} \otimes c_{t-1} + b_\text{f})\\
c_{t}=& f_{t} \otimes c_{t-1} + i_{t} \otimes \tanh(W_\text{c}*z_{t}+V_\text{c} * h_{t-1} + b_\text{c})\\
o_{t}=& \sigma(W_\text{o}*z_{t} + V_\text{o}*h_{t-1} + U_\text{o} \otimes c_{t} + b_\text{o})\\
h_{t}=& o_{t} \otimes \tanh(c_{t}), 
\end{aligned}
\end{equation}
where $*$ and $\otimes$ are the convolution operations and the element-wise product, respectively; $\sigma$ and $\tanh$ are the sigmoid and hyperbolic tangent non-linearities. $i_t$, $f_t$, and $o_t$ are the input, forget, and output gates, respectively; $c_t$ and $h_t$ are the cell and the latent state, where $g = h_T$. $W_l$ and $V_l$ for $l \in \{\text{i}, \text{f}, \text{c}, \text{o}\}$ are trainable convolution kernels; $U_l$ and $b_l$ are trainable parameters of the same size as $z_t$. Multiple ConvLSTM can be stacked and temporally concatenated to form more complex structures and may further improve performance. In our network, we only use a single layer ConvLSTM.

\begin{table*}[t]
\centering
\begin{tabular}{lccc}
\toprule
\multirow{2}{*}{Models}&\multicolumn{2}{c}{Cityscapes}&\multicolumn{1}{c}{EndoVis2018}\\
\cmidrule(lr){2-3}
&Validation& Test& Validation\\
\midrule
\midrule
FCN-8s \cite{long2015fully} &64.3 &- & 47.9\\
DeepLab-v3 \cite{chen2017rethinking} &71.8 &- & 56.2\\
DANet \cite{fu2019dual} &68.7 &- &56.0\\
PSPNet (baseline) \cite{zhao2017pyramid} &71.6 &71.0 &59.8\\
ICNet (baseline) \cite{zhao2018icnet} &60.0 &59.5 &52.1\\
\midrule
DynamicCRF \cite{b8bbc70efa1c4193927e1538a11d177f} &64.5 &- &-\\
ConvLSTM \cite{pfeuffer2019semantic} &62.3 &- &-\\
GRFP \cite{nilsson2018semantic} &\textbf{73.6} &\textbf{72.8} &-\\
\Mname{} (ICNet) (\textit{w.o.} noisy-training) &61.2 &60.5 &53.6\\
\Mname{} (ICNet) &62.5 &61.6 &54.8\\
\Mname{} (PSPNet) (\textit{w.o.} noisy-training) &72.2 &71.7 & 61.1\\
\Mname{} (PSPNet) &73.0 &\textbf{72.8} & \textbf{62.3}\\
\bottomrule
\end{tabular}
\caption{Comparison results with the state-of-the-art methods on Cityscapes and EndoVis datasets.
All predictions are evaluated with mIoU(\%). Best performance in bold.}
\label{tab1}
\end{table*}

\subsection{Enhancing Temporal Awareness}\label{noise}
For video tasks, the temporal coherency between frames is often leveraged for better performance. However, there might be some cases in which this affects negatively. For example, in surgery videos, consecutive frames may usually have small motions and occasionally exhibit large motions. Such rare events may not be well learned with, \eg, RNN-based models.

For neural network training, a number of attempts have been made to utilize noises in various ways for the sake of regularization \cite{noh2017regularizing,bishop1995training}. More recently, some studies demonstrated that huge amount of unlabeled data, which may serve as noises in training, can improve the performance of semantic segmentation and classification tasks in teacher-student networks \cite{xie2018improving,xie2019self}. Inspired by these works, we propose a noisy-training strategy, which replaces some frames in input sequences with unlabeled and random images during training. This noise injection in the time domain stochastically spoils temporal dependency in the original sequence and may consequently improves the capability of feature extraction from individual frames as the temporal continuity is no longer reliable, and thus we can expect better temporal awareness in the model.

Specifically, for each sequence, we replace some of context frames with random frames, which are unlabeled random images with much different contents, as shown in Fig.~\ref{fig:noise_pic}. For example, we may use  handwriting images, frames in TV drama series, or medical images as noise to replace frames when dealing with street-view sequences. Even random tensor can be used as one type of noise. The target frame are not replaced, so that we can still use its ground-truth label. In addition, due to the structural characteristics of our model, the feature maps from context frames are used solely for enhancing the target frame's feature map, and the output from the model is the segmentation result for the target frame. This means that there is no need to generate, \eg, pseudo labels for noises, which are required in \cite{xie2018improving,xie2019self}. Therefore, the noisy-training strategy causes no extra computation nor annotation. 

For adding noise, each context frame (\ie, $s_1, \dots, s_{T-1}$) is replaced with  a random image with the probability of $p$, which is set to 50\% in our implementation. We also limit the number of frames to be replaced to half of sequence length (\ie, $T/2$). It means replaced frames is no more than two in our experiment. 
In addition, we introduce another kind of temporal noises, \ie{}, randomly reversing the previous frames in input sequences. This operation also helps interupt temporal data.


\begin{figure*}[t]
\centering 
\includegraphics[width=0.9\textwidth]{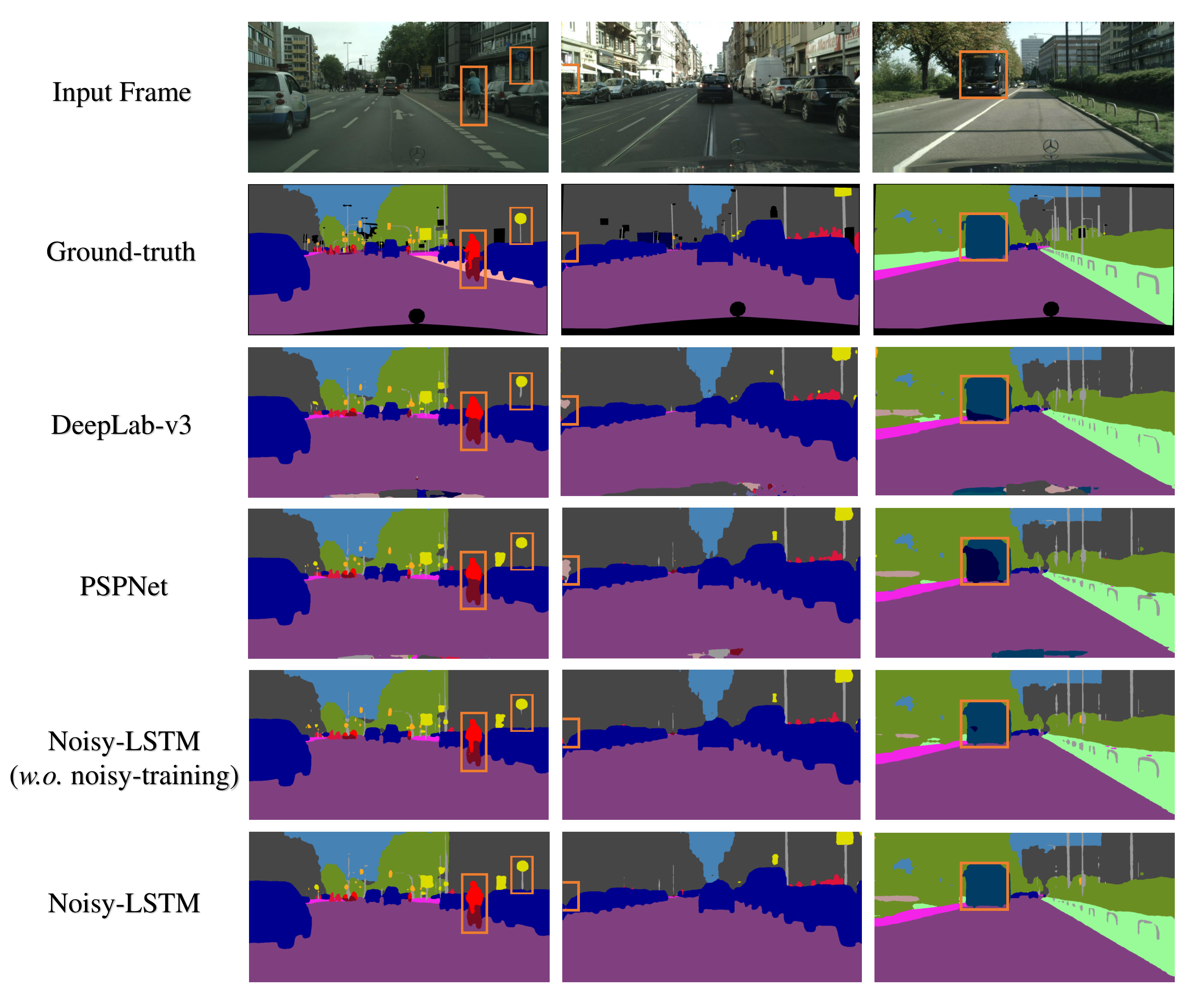}
\caption{Visualization of the segmentation results of the Cityscapes dataset using PSPNet based model. Observe how ConvLSTM plus noise strategy is able to correct wrong segmentation by favorable obtaining previous frames information.
Notable differences are marked with orange boxes.}
\label{fig:city_compare} 
\end{figure*}

\begin{figure*}[t]
\centering 
\includegraphics[width=1\textwidth]{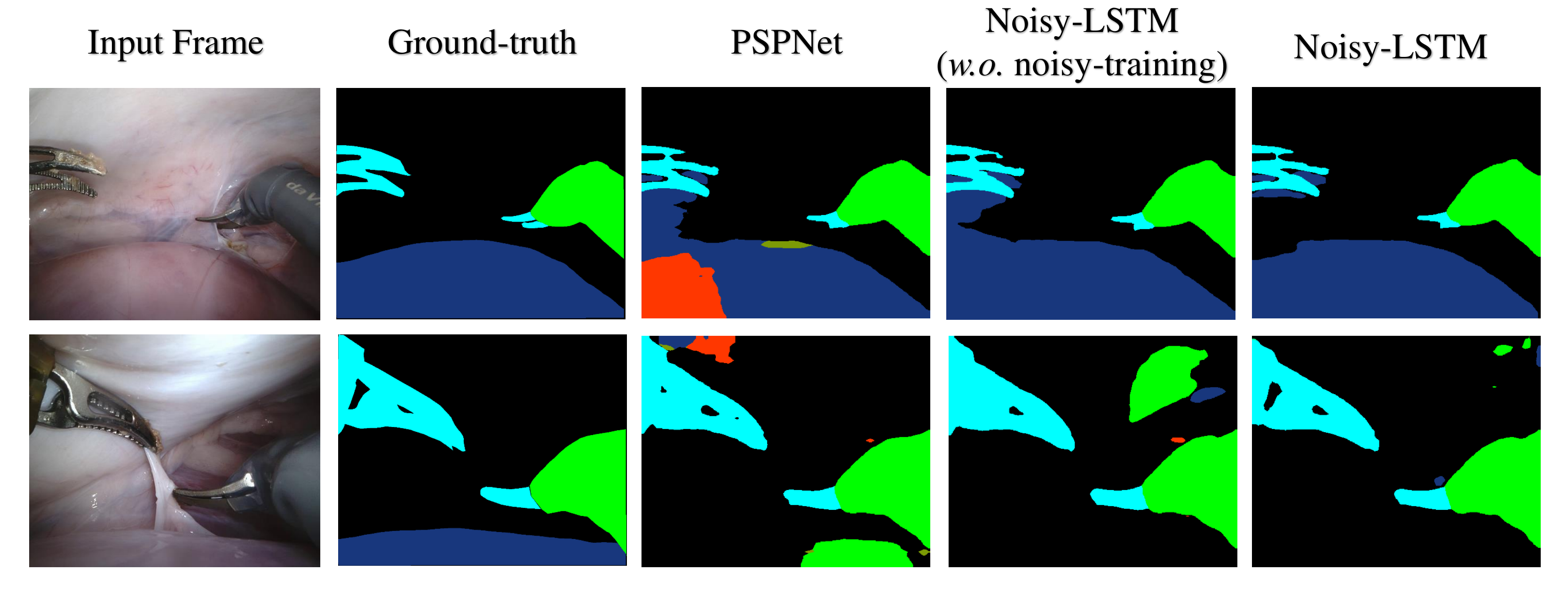}
\caption{Visualization results from the EndoVis2018 dataset using PSPNet based model. The \mname{} model can get more accurate segmentation result on the body tissues in the first row, and the surgical instruments in the second row.}
\label{fig:miccai_compare} 
\end{figure*}

\section{Experiments}
In order to evaluate our model trained with the noisy-training strategy, we used two video semantic segmentation datasets in completely different domains, \ie, Cityscapes \cite{cordts2016cityscapes} and EndoVis2018 \cite{allan20202018}. Frames in one dataset were used as noises when training our model for the other dataset, whereas labels in the dataset used as noises were not used in this process. For data augmentation to all experiments, we adopted operations including rotate (angle between -10 and 10), random horizontal flipping, and so on. When training with temporal data, all the input images in one sequence will calculated by the same data augmentation.  

We used cross-entropy as the loss function and Adam \cite{kingma2014adam} as the optimizer with an initial learning rate of $10^{-4}$, which was decreased with the factor of 10 after half way of the training. The iteration was terminated after 40 epochs for Cityscapes and 30 epoches for EndoVis2018.
The length $T$ of sequence was set to 4, and the number $N$ of sequences was also set to $4$. The hidden state $h_0$ and cell $c_0$ were zero initialized. The model was implemented in Pytorch\cite{paszke2017automatic} framework and we ran the model on the Tesla V100 GPU with 32GB memory.

\subsection{Cityscapes Dataset}\label{city}
The Cityscapes dataset contains in total 5,000 video sequences of high-resolution frames ($2,048 \times 1,024$), partitioned into training, validation, and test sets with 2,975, 500, and 1,525 sequences, respectively. The videos are captured in different weather conditions across 50 different cities in Germany and Switzerland. There are $30$ categories in total in the Cityscapes dataset, however, follow the previous research, only $19$ of them are used in the semantic segmentation task.

We try with different lengths of the frame interval and find that we can achieve the best performance with an interval of $0.12$s (more details in Sec. \ref{skip_section}), which is adopted for all methods. Because of the GPU memory issues caused by high-resolution images in Cityscapes and a larger visual field, we resize the original image into $1024\times512$ for PSPNet-based model and apply a sliding window with a size of $448\times448$ on the resized images. For ICNet-based model, we maintain the original resolution and adopt sliding window with a size of $512\times1024$. We firstly train the network without ConvLSTM module for 40 epochs. After that, the whole network is trained for another 40 epochs. The results of the best performing model on the validation set are submited to the Cityscapes test server.

The results are summarized in Table \ref{tab1}. For comparison, we evaluated FCN-8s \cite{long2015fully}, DeepLab-v3 \cite{chen2017rethinking}, and DANet \cite{fu2019dual}, all of which were re-implemented and trained in the same configuration. We also recorded the results reported by previous research. It turned out that, compared to the baseline model, PSPNet-based (ICNet-based) \mname{} achieved better performance, with improvements of 1.4\% (2.5\%) in the validation set and 1.8\% (2.1\%) in the test set. In addition, the noisy-training strategy also improves the performance in both validation and test sets. We also show some qualitative results from the validation set in Fig.~\ref{fig:city_compare}, Every column lists the input image with its ground true label and model prediction. All the notable changes are highlighted in orange boxes. 
It shows that \Mname{} can generate accurate predictions on some challenging objects. For example, the human body in the first column (marked in red), the wall in the second column, and the bus in the third column. Actually, all these objects exist in the previous frames. We can see that \Mname{} can obtain information from these frames and fix wrong segmentation. In this case, the noisy-training strategy can help the network to obtain these kinds of temporal information more efficiently.

\subsection{EndoVis2018 Dataset}\label{EndoVis}

We also evaluated and compared \mname{} on the EndoVis2018 dataset \cite{allan20202018}. EndoVis2018 dataset includes 19 sequences, which is split into 15 and 4 sequences for training and testing. We picked up two sequences (sequences \#5 and \#10) from the training set and used them as the validation set. We resized the image into $520\times416$ for PSPNet-based model (ICNet-based model use original resolution as input) during training and recovered it into original resolution for evaluation. Each pixel in the frames are annotated with one of 11 class labels, including organ tissues and surgical instruments.

Table \ref{tab1} shows that \mname{} model can also outperform other methods on this dataset. Some examples are present in Fig.~\ref{fig:miccai_compare}. Similarly to the Cityscapes dataset, our \mname{} gives accurate segmentation even of small regions.

\subsection{Effects of Hyperparameter}\label{Extra}\label{skip_section}
There are some important parameters related to the network performance. This section gives some extra experimental results to show the effect of the frame interval and the number of input sequences over the Cityscapes dataset's validation set.

\paragraph{Frame interval}
For Cityscapes, each video sequence has 30 frames at 16.7 fps, and the 20-th frame was annotated. \Mname{} model contextualizes the target frame with $T-1$ precedence frames, and context frames can be chosen arbitrarily. In our implementation, we re-sample the context frames from the video sequence, \ie, there are a constant number of frames in-between $s_t$ and $s_{t+1}$. We evaluated the cases when context frames are sampled every 1, 2, and 5 frames, which corresponds to frame intervals of 0.12s, 0.18s, and 0.36s, respectively.
Table \ref{tab_skip} shows the results of the proposed model with or without noisy-training using different frame intervals.
The best result is obtained with a interval of $0.12$s and with noisy-training strategy.
It shows that longer interval leads to the decrease of the segmentation performance. 
Also, in all temporal intervals, noisy-training methods can always show the correction capability. This fact proves that the noisy-training strategy will enhance the temporal awareness of the deep learning models and give them a better ability to extract useful information among previous frames.

\paragraph{Type and probability of noises}
Noisy-training is the key to this research. Thus, we evaluate the effects of different types and intensity of the added noises with PSPNet-based model. We add three different types of noises including unrelated data, random tensor, and extreme augmentation (distortion or Gaussian blur). For the noise intensity which means the probability of noise appears in previous frames, we use 25\%, 50\%, 75\%, and 100\%. The result shows that both unrelated data and random initialization can improve the prediction, while extreme augmentation can not perform as an ideal type of noise. For noises of unrelated data, probability does not obviously affect the performance and the best result is obtained with the probability of 50\%. For the noise type of random tensor, the increase of noise probability will deteriorate the model's performance.

\begin{table}[t]
\centering
\resizebox{3.35in}{!}{
\begin{tabular}{ccc}
\toprule
\multirow{2}{*}{Frame Interval(s)}&\multicolumn{2}{c}{mIoU(\%)}\\
\cmidrule{2-3}
&\textit{w.o.} Noisy-training &
\textit{w.} Noisy-training\\
\midrule
0.06& 72.0 &72.7 \\
0.12& 72.2 &\textbf{73.0} \\ 
0.18& 71.6 &72.6\\
0.36& 71.2 &71.9\\
\bottomrule
\end{tabular}
}
\caption{Results with different intervals between input frames. All experiments use PSPNet as base network and evaluated in cityscapes validation set. Best result is obtained by skip distance 2 using noise strategy.
}
\label{tab_skip}
\end{table}

\begin{table*}[t]
\centering
\begin{tabular}{ccccc}
\toprule
\multirow{2}{*}{Probability}&\multicolumn{4}{c}{Noise Type}\\
\cmidrule{2-5}
&Unrelated Data &Random &Distortion &Gaussian Blur \\
\midrule
0\% & 72.2 &72.2 &72.2 &72.2\\
25\%& 72.4 &72.5 &71.6 &71.4\\ 
50\%& \textbf{73.0} &72.9 &71.2 &71.7\\
75\%& 72.8 &72.4 &71.0 &71.5\\
100\%& 72.5 &71.0 &71.2 &71.3\\
\bottomrule
\end{tabular}
\caption{Results with different noise type and probability. All experiments use PSPNet as base network and evaluated in cityscapes validation set. Best result is obtained by noise type with unrelated data and probability 50\%.
}
\label{tab_noise}
\end{table*}

\begin{table*}[t]
\centering
\begin{tabular}{lcccccc}
\toprule
\# GPUs      & \multicolumn{4}{c}{1} &  \multicolumn{2}{c}{2} \\ 
\cmidrule(lr){2-5} \cmidrule(lr){6-7}
\# sequences&1 &  2   &  3   &  4   &  4   &  8 \\ \midrule
ICNet-based&61.3 &61.9 &61.7 &62.5 &61.8 &62.0 \\
PSPNet-based&68.4 & 71.9 & 72.4 & 73.0 & 72.6 & 73.3 \\
\bottomrule
\end{tabular}
\caption{Results with different numbers of sequence batches in Cityscapes validation set.}
\label{tab_batch}
\end{table*}

\begin{table*}[t]
\centering
\begin{tabular}{lccc}
\toprule
\multirow{2}{*}{Model}&\multicolumn{3}{c}{Noise Type}\\
\cmidrule{2-4}
&Distortion& Gaussian Blur& No noise\\
\midrule
ICNet-based (\textit{w.o.} noisy-training) & 57.5 &58.8 &61.2 \\ 
ICNet-based & 61.7 &62.0 &62.5 \\
PSPNet-based (\textit{w.o.} noisy-training) & 70.8& 71.1&72.2\\
PSPNet-based & 72.6& 72.4&73.0 \\
\bottomrule
\end{tabular}
\caption{Results of anti-noise experiment.}
\label{tab_anti}
\end{table*}

\paragraph{Number of input sequences in a batch}
This parameter is critical for BN and thus can affect the performance. We evaluate the effect of the sequence numbers in both PSPNet-based and ICNet-based Noisy-LSTM models. We train models with different sequence numbers on single or multiply GPUs and the results are shown in Table.~\ref{tab_batch}. We can see that, in PSPNet-based \mname{}, a larger batch size lead to better prediction performance. When the batch is set to $1$, it will cause a great performance drop. For ICNet-based \mname{}, improving the batch size can slightly improve the performance. The experimental results prove the necessity of bn for training.

\begin{figure*}[t]
\centering 
\includegraphics[width=1\textwidth]{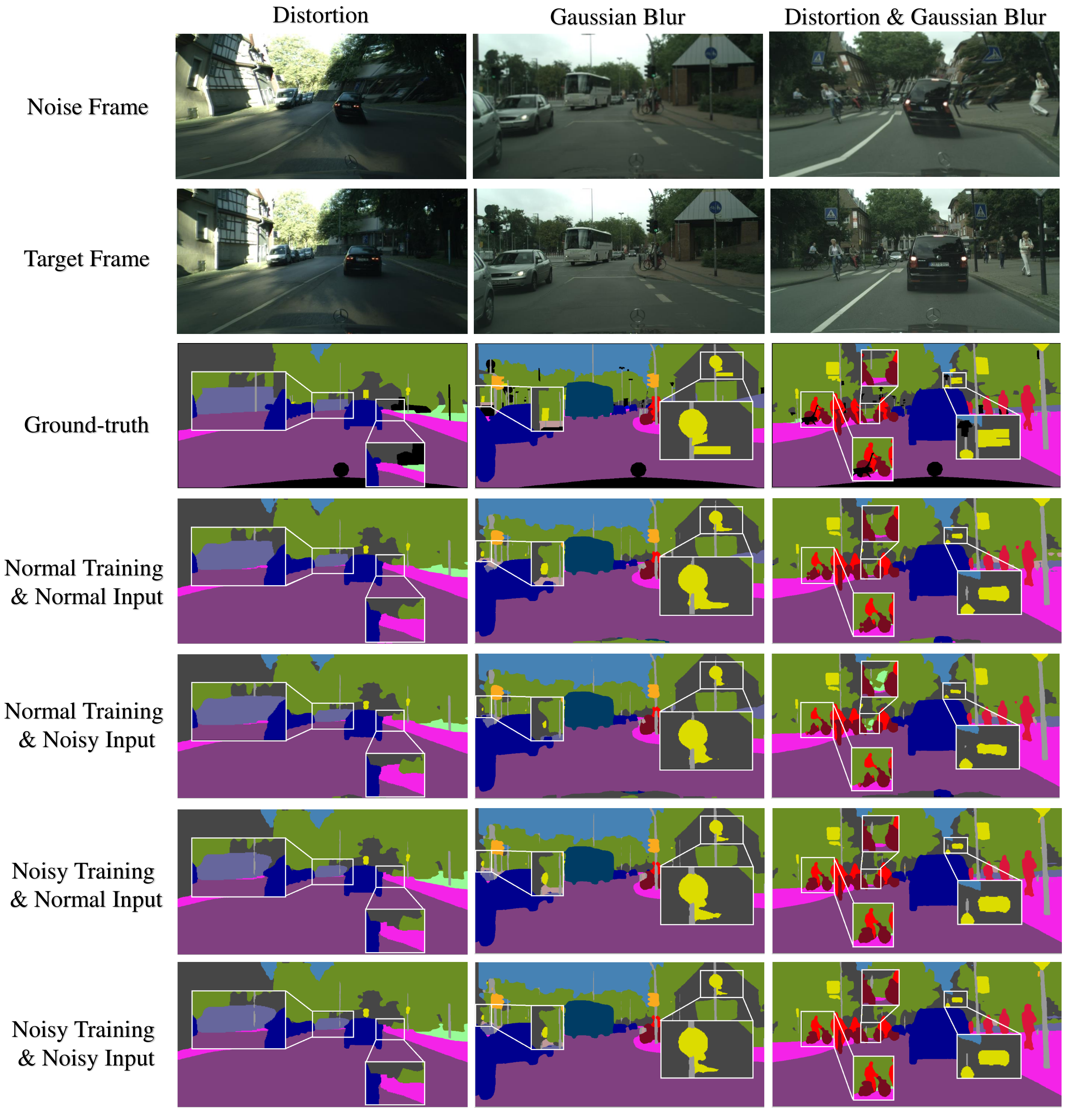}
\caption{Visualization results of anti-noise experiment.}
\label{fig:anti_compare} 
\end{figure*}

\paragraph{Anti-noise experiment}
For video tasks, when the image quality of the previous frame is not good due to external factors (blurring, \textit{etc.}), much noisy information is included in the temporal features. In this case, the prediction of the target frame will be affected and the performance may decrease. Our \mname{} can overcome this problem and generate accurate segmentation masks in some extreme situations. In Table.~\ref{tab_anti}, we show the results of the anti-noise experiment for both ICNet-based and PSPNet-based model in the validation set of the Cityscapes dataset. We applied two kinds of noises (Gaussian blur and distortion) and added them to the first and third frames in the input sequence for evaluation. The noisy images are shown in Fig.~\ref{fig:anti_compare}. We found that normally-trained models will be influenced by the noisy input, while the noisy-training strategy can weaken this performance degradation. We also provide some comparison samples in Fig.~\ref{fig:anti_compare} (obvious differences are marked with magnifying glasses). All these results are generated by the PSPNet-based model. The target frame is the last frame of the continuous four frames in the input sequence and the noise frame is the third frame added with noises. In the first column, we applied distortion to the noise frame; in the second column, we adopted Gaussian blurring; in the third column, we applied both distortion and Gaussian blurring.
Compared to the normal input results, noisy input will cause the wrong predictions. For example, in the first column, the predictions of the wall (slate blue) should be the building (grey). Also, the end of the sidewalk (fuchsia) also covered the wrong areas. We think this phenomenon is due to the distortion of objects on the previous frame, which conveys these incorrect temporal information to the prediction for the target frame. On the contrary, the results of noisy-trained models are only slightly affected. Similarly, in the second column, Gaussian blurring will blur the outline of small objects and even blend them into the surrounding environment. With noisy input, normally-trained models will give incomplete predictions of signboard (yellow), while noisy-trained models have a much better performance.

\section{Conclusion}
In this paper, we proposed a model named \mname{} for semantic video segmentation, which is trainable in an end-to-end manner. \Mname{} is capable of utilizing the temporal dependency in video sequences to improve the segmentation performance. It employs a single layer convolutional LSTM to encode spatio-temporal features. In addition, we propose the noisy-training strategy, which introduces noises during training so as to avoid excessive reliance on precedence frames and thus is expected to improve feature extraction. Our experimental results demonstrated that this strategy further improved the performance without extra data annotation or computational costs, achieving the state-of-the-art performances on the Cityscapes and EndoVis2018 datasets.

\begin{acknowledgements}
This work was supported by Council for Science, Technology and Innovation (CSTI), cross-ministerial Strategic Innovation Promotion Program (SIP), ``Innovative AI Hospital System'' (Funding Agency: National Institute of Biomedical Innovation, Health and Nutrition (NIBIOHN)). This work was also supported by JSPS KAKENHI Grant Number 19K10662 and 20K23343.
\end{acknowledgements}

%
%


\bibliography{mybibfile}

\end{document}